\documentclass[10pt,twocolumn,letterpaper]{article}

\usepackage[pagenumbers]{cvpr} 

\usepackage{graphicx}
\usepackage{lipsum} 
\usepackage{pifont}
\usepackage{multirow}
\usepackage{array}
\usepackage{booktabs}
\definecolor{cvprblue}{rgb}{0.21,0.49,0.74}
\usepackage[pagebackref,breaklinks,colorlinks,allcolors=cvprblue]{hyperref}

\title{ZDySS - Zero-Shot Dynamic Scene Stylization using Gaussian Splatting}

\author{Abhishek Saroha$^{1,2}$~\quad Florian Hofherr$^{1}$~\quad Mariia Gladkova$^{1}$\\ Cecilia Curreli$^{1}$  ~\quad Or Litany$^{3,4}$~\quad Daniel Cremers$^{1,2}$
\\[0.2ex]
\small{$^{1}$Technical University of Munich} \quad 
\small{$^{2}$Munich Center for Machine Learning} \quad 
\small{$^{3}$Technion} \quad 
\small{$^{4}$Nvidia}\\
}

\begin{document}

\twocolumn[{
\maketitle
\begin{center}
    \captionsetup{type=figure}
    \includegraphics[width=\textwidth, height=6cm]{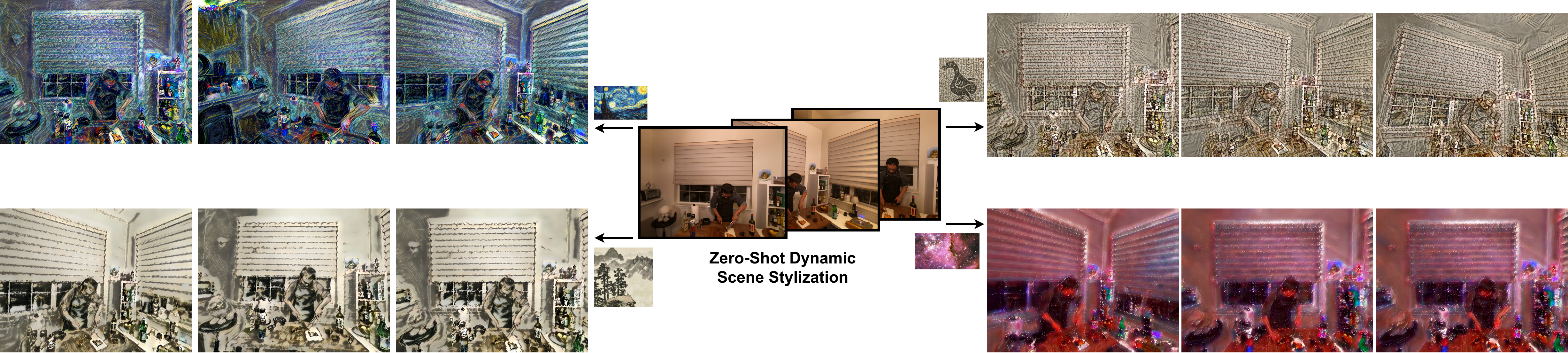}
    \caption{Given a multi-view video of a dynamic scene, we present a method that once trained on  the scene, is able to perform high quality stylization from unseen style images across novel views and timesteps while maintaining consistency in the spatio-temporal domain.}
    \label{fig:teaser}
\end{center}
}]

\begin{abstract}

Stylizing a dynamic scene based on an exemplar image is critical for various real-world applications, including gaming, filmmaking, and augmented and virtual reality. However, achieving consistent stylization across both spatial and temporal dimensions remains a significant challenge. Most existing methods are designed for static scenes and often require an optimization process for each style image, limiting their adaptability. We introduce ZDySS, a zero-shot stylization framework for dynamic scenes, allowing our model to generalize to previously unseen style images at inference. Our approach employs Gaussian splatting for scene representation, linking each Gaussian to a learned feature vector that renders a feature map for any given view and timestamp. By applying style transfer on the learned feature vectors instead of the rendered feature map, we enhance spatio-temporal consistency across frames. Our method demonstrates superior performance and coherence over state-of-the-art baselines in tests on real-world dynamic scenes, making it a robust solution for practical applications.

\end{abstract}    
\section{Introduction}
\label{sec:intro}

Art, in all various forms, has been instrumental in captivating human creativity. Artworks, especially in the forms of paintings, have given humans diverse insights into the lives, culture and perspectives of various people across different eras. Bridging the gap between creativity and reality, the inspiring work of Gatys \etal \cite{gatys2016image} proposed to use neural networks for transferring the style of an artwork to any real image. With the advent of abundant 3D data, and enhancements in reconstruction techniques such as NeRFs \cite{mildenhall2021nerf} or 3D Gaussian Splatting (3DGS)~\cite{kerbl3Dgaussians}, the concept of neural style transfer was extended to stylize entire scenes ~\cite{liu2023stylerf, zhang2022arf, huang2022stylizednerf, nguyenphuoc2022snerf, zhang2023refnpr}. However, these approaches primarily focus on static data, which does not fully capture the dynamic nature of real-world scenes.
In this paper, we address the challenge of stylizing dynamic scenes, an important step towards more accurate and immersive style transfer applications depicting the world around us in motion.

The task of scene editing is crucial to many real-life use cases, ranging from games, movies, all the way to augmented and virtual reality applications. Another important application for such tasks is modeling and modifying digital avatars, as demonstrated in the works of ~\cite{qian2023gaussianavatars,pang2023ash,moreau2023human, zheng2023gpsgaussian} to name a few. Therefore, it is required that not only such stylization methods are efficient, but rather flexible with the type of styles they deal with. To this end, we focus on a zero-shot method for stylizing dynamic scenes, that ensures that once a model is trained on a particular dynamic scene, it does not need any further optimization during test time for any queried style. Most prior works, in static ~\cite{zhang2022arf,zhang2023refnpr,nguyenphuoc2022snerf,jung2024geometry} and dynamic scenes ~\cite{li2024sdyrf} are based on this setting, thereby limiting their practical applicability. 

Most such stylization methods are developed for Neural Radiance Fields (NeRF)~\cite{mildenhall2021nerf}. The simplicity of representing a scene using the weights of a learnt neural network, an extremely simple multi-layer perceptron (MLP) quickly established NeRFs as the favorite tool to solve the problem of Novel View Synthesis(NVS). Despite being initially computationally expensive, more recent follow-up works such as ~\cite{reiser2021kilonerf,muller2022instant} and ~\cite{barron2021mipnerf,barron2023zipNerf} among others focused on speed and quality improvements respectively. More recently, 3D Gaussian Splatting(3DGS)~\cite{kerbl3Dgaussians} developed a novel pipeline for NVS, using a more explicit representation leveraging blob-like structures, termed as 3D Gaussians, that were, at par, if not better than NeRFs while also being faster during training and inference. While 3DGS focused on static scenes, works such as ~\cite{yang2024deformable,wu20244dgs} to name a few, extended the framework to incorporate dynamic scenes. In this work, we also build up on the backbone of ~\cite{wu20244dgs}.

The task of stylizing dynamic scenes presents several challenges, with the primary difficulty being maintaining consistency across multiple views while ensuring temporal coherence. In this context, the problem setup involves training with paired camera poses and images from a specific time frame to generate stylized novel views in either the spatial or temporal domain, conditioned on a given style image. Only a few works, such as ~\cite{li2024sdyrf} and ~\cite{xu2024styledyrf}, have addressed the task of dynamic scene stylization. The method of Li \etal \cite{li2024sdyrf} requires costly optimization for each queried style image. In contrast, the approach by Xu \etal \cite{xu2024styledyrf} offers a zero-shot solution,
however, it depends on an MLP-based stylization transformation, which requires a large dataset for training.

In this work, we present ZDySS, a novel end-to-end trainable stylization pipeline for dynamic scenes based on Gaussian splatting. We use previous work to enhance each Gaussian by a feature vector, which allows us to lift 2D VGG features to the 3D space. These features allow us to adapt the well-known 2D stylization approach of Adaptive Instance Normalization (AdaIN) \cite{huang2017adain} to our pipeline. We propose to rely on a running average to ensure spatial and temporal consistency of the volumetric feature statistics. Once a dynamic Gaussian scene is trained, our method offers zero-shot stylization with arbitrary styles. The results show dynamic scene stylizations across a variety of styles, achieving compelling visual effects.

In brief, the contributions of our paper are as follows:

\begin{itemize}
    \item We present ZDySS a novel end-to-end trainable stylization pipeline for dynamic scenes based on Gaussian splatting for which we adapt Adaptive Instance Normalization (AdaIN) \cite{huang2017adain} to the spatial-temporal domain.
    \item In contrast to previous work, we do not need a pre-trained style transfer module.
    \item Unlike common existing paradigms, we operate in a zero-shot manner, i.e. do not need any training or test time optimization on the queried style image and can handle arbitrary, unseen style images while maintaining temporal and spatial consistency. 
\end{itemize}

\noindent To ensure the reproducibility of this work, we give additional details and experiments in the supplementary material and will release our code upon acceptance. Please also see our supplementary videos.

\begin{table}[h]
\centering
\label{tab:overview}
\small
\begin{tabular}{c|c|c|c}
\hline
Method    & {End-to-End Training}  & \#Styles & Input\\ 
\hline \hline
4DGS~\cite{wu20244dgs} & \textcolor{green}{\ding{51}}  & $ 1 $ & Multi-View\\
\hline
StyleDyRF\cite{xu2024styledyrf} & \textcolor{red}{\ding{55}}  & $ \infty$ & Monocular\\
\hline
S-Dyrf~\cite{li2024sdyrf} & \textcolor{red}{\ding{55}}  & $ 1$ & Multi-View\\
\hline \hline
ZDySS(Ours)   & \textcolor{green}{\ding{51}} & $ \infty$ & Multi-View\\ 
\hline
\end{tabular}
\caption{
\textbf{Methods Overview} A comparison overview of the different methods and their salient features. 
In contrast to other works, our approach offers both end-to-end training and the ability to stylize a scene with arbitrary styles in a zero-shot manner at inference.
}
\end{table}
\section{Related Works}
\label{sec:formatting}

\subsection{Scene Representation with Radiance Fields}
There has been an explosion in work on scene representation based on radiance fields, which can be attributed to NeRF-like methods \cite{mildenhall2021nerf} and Gaussian splatting-based approaches \cite{kerbl3Dgaussians}. We review the most relevant work for radiance field-based static and dynamic scene representation in the following.


\paragraph{Static Scenes}

Building on the seminal work by Mildenhall et al. \cite{mildenhall2021nerf}, \emph{neural} radiance fields (NeRFs) combine a multilayer perceptron (MLP) representing color and density with volumetric rendering to enable highly realistic scene reconstructions. 
Countless improvements have been proposed, addressing inference speed \cite{reiser2021kilonerf,lindell2021autoint}, large-scale reconstruction \cite{tancik2022blockNerf}, anti-aliasing \cite{barron2021mipnerf,barron2022mipNerf360}, appearance changes \cite{martin2021nerfw,tancik2022blockNerf}, and many more.
One notable direction of research aims to speed up NeRFs by employing learnable features in spatial structures like planes \cite{fridovich2023kPlanes} or grids \cite{muller2022instant, barron2023zipNerf}, which allows the use of significantly smaller MLPs.
Other works using spatial feature structures completely eliminate the need for MLPs, marking a shift toward non-neural radiance field representations \cite{yu2021plenoxels, Chen2022tensorf}.

All previous methods employ \emph{volume rendering} to generate images from the radiance field. However, this requires inefficient stochastic sampling, which is time-consuming and can lead to noise. In contrast, 3D Gaussian splitting (3DGS) \cite{kerbl3Dgaussians} shows state-of-the-art real-time radiance field rendering by using a \emph{rasterization-based rendering} approach. 


\paragraph{Dynamic Scenes}
The work on dynamic NeRFs can be classified into two categories: A first class of approaches constructs a time-dependent NeRF by adding an additional input dimension \cite{li2021neuralSceneFlowFields,xian2021space,gao2021dynamic,Cao2023HexPlane,fridovich2023kPlanes} for the time. While this approach is elegant, multiple additional loss terms and regularizers are usually required to disentangle the scene's static and dynamic parts and ensure temporal consistency.
In contrast, several other works combine a scene representation in a reference configuration with a time-dependent deformation field to represent the dynamic scene \cite{park2021nerfies, park2021hypernerf,pumarola2021dNerf,tretschk2021nonRigidNeuralRadianceFields,wu2022d2Nerf}. Inherently, these approaches struggle with changing scene topology like appearing objects.
Most of the dynamic Gaussian splatting methods use a canonical space and a displacement field \cite{luiten2024dynamic,yang2024deformable,wu20244dgs}. Other works explore  enhancing 3D Gaussians by temporal attributes \cite{li2024spacetime} or temporal slicing of 4D Gaussians \cite{duan20244d,yang2023real}.


\subsection{Style Transfer}
Style transfer is the task of reimagining an image or a video in the style of a reference image. In one of the pioneering works, Gatys \etal \cite{gatys2016image} introduced an optimization-based approach for image style transfer using features from pre-trained CNNs. To eliminate the need for costly optimization per image, Johnson \etal \cite{johnson2016perceptual} employed perceptual losses to train feed-feed forward networks capable of applying a fixed style to an arbitrary image. Finally, adaptive instance normalization (AdaIN) \cite{huang2017adain} enabled real-time style transfer from arbitrary sources by using the AdaIn layer as a feature transformation in combination with an encoder-decoder pair. Several follow-up works have explored alternative transformations for this architecture, including carefully crafted whitening and coloring transforms \cite{li2017universalStyleTransferFeatureTransforms}, multi-scale style decorators \cite{sheng2018avatarNet}, transformations learned from data \cite{li2019learningLinearTransformsStyleTransfer}, and attention-based approaches \cite{deng2020arbitrary,liu2021adaattn,park2019arbitraryStyleTransferStyleAttentionalNets,wu2021styleformer}. Svoboda \etal \cite{svoboda2020twoStage} propose a graph convolutional transformation layer and moreover eliminate the need for a pre-trained network for the perceptual loss by introducing a set of cyclic losses.

Style transfer for video imposes the additional challenge of temporal consistency to avoid flickering artifacts. While several works rely on optical flow \cite{chen2017coherent,huang2017real,ruder2018artisticStyleTransferVideos}, others use regularization-based approaches \cite{wang2020consistent,wang2020consistentVideoStyleCompoundReg} or use stylized keyframes \cite{jamrivska2019stylizingVideoByExample} that propagate information to the video.


\subsection{Scene Stylization} 

Scene stylization is the process of applying a style to a static or dynamic 3D scene to enable visually appealing novel view synthesis. One major challenge in this task is to ensure multi-view consistency to avoid flickering artifacts.

\paragraph{Static Scene Stylization}
While some earlier works have utilized classical data structures like point clouds \cite{huang2021learningToStylizeNovelViews,mu20223d} or meshes \cite{hollein2022stylemesh} for static scene stylization, the focus has shifted to radiance field-based scene representations. One approach is to adapt a pre-trained NeRF to match a given style. Nguyen-Phuoc \etal \cite{nguyenphuoc2022snerf} use individually stylized images rendered from a pre-trained NeRF as targets, iteratively fine-tuning the model to align with the desired style in a view-consistent manner. Artistic Radiance Fields (ARF) \cite{zhang2022arf} follows a similar approach but employs a nearest neighbor matching loss on the features extracted from rendering and style instead of a standard color loss. Pang \etal \cite{pang2023locally} propose an adaption-based approach that adapts style patterns better onto local regions, while Zhang \etal \cite{zhang2023refnpr} use a reference ray registration strategy to reduce the number of required stylized reference images significantly.

The refinement of the NeRF in all of these methods is costly. To allow for zero-shot style transfer, other methods aim to modify the color module of a pre-trained NeRF based on a style code extracted from a single style image. While Chen \etal \cite{chen2022upstnerf} and Chiang \etal \cite{chiang2022stylizing} employ a hypernetwork to modify the color module, Huang \etal \cite{huang2022stylizednerf} completely replace it with a predicted style network.
Liu \etal \cite{liu2023stylerf} train a NeRF with features in combination with a decoder and apply a style transformation to the rendered feature map.
Other works explore stylization for neural SDFs \cite{fan2022unified} or stylization based on a text prompt using CLIP embeddings \cite{wang2023nerfArt}.

Recent work also explores the stylization of Gaussian splatting-based scenes. Mei \etal \cite{mei2024referencegs} refine a pre-trained 3DGS to a single style using a texture-guided control mechanism. Kovács \etal \cite{kovacs2024gsplat} augment this idea with a pre-processing step. By employing a color module that modulates the color of the Gaussians based on position and the output of a style encoder, Saroha \etal can stylize a pre-trained 3DGS to an arbitrary style.
Liu \etal \cite{liu2024stylegaussian} embed VGG features onto the Gaussians and use an AdaIN layer for the style transfer.
While this approach is similar in spirit to ours, they only consider \emph{static} scenes. Moreover, they use a 3D RGB decoder which necessitates a large style dataset for pre-training.


\paragraph{Dynamic Scene Stylization}
Applying stylization to dynamic scenes is a very new field of research addressed in only a few works. Li \etal \cite{li2024sdyrf} refine a pre-trained dynamic NeRF using pseudo-references, which are created by transferring the style of a stylized reference frame to an entire rendered reference video through a video style transfer method. Compared to our approach, this method needs to be re-trained for each style.
Similar to our work Xu \etal \cite{xu2024styledyrf} also use volumetric features to augment their NeRF-based dynamics model and use a decoder to obtain RGB values from rendered features. In contrast to our work, however, they use a learned transformation MLP for stylization, which necessitates a large style dataset. Moreover, training their NeRF-based dynamic scene model requires significantly more time than our approach.
The concurrent preprint by Liang \etal \cite{liang20244dstylegaussianzeroshot4dstyle} adopts this idea to dynamic Gaussian splatting. In contrast to our work, they do not employ a running average on the feature statistics.

\begin{figure*}[h]
    \centering
    \includegraphics[width=\textwidth]{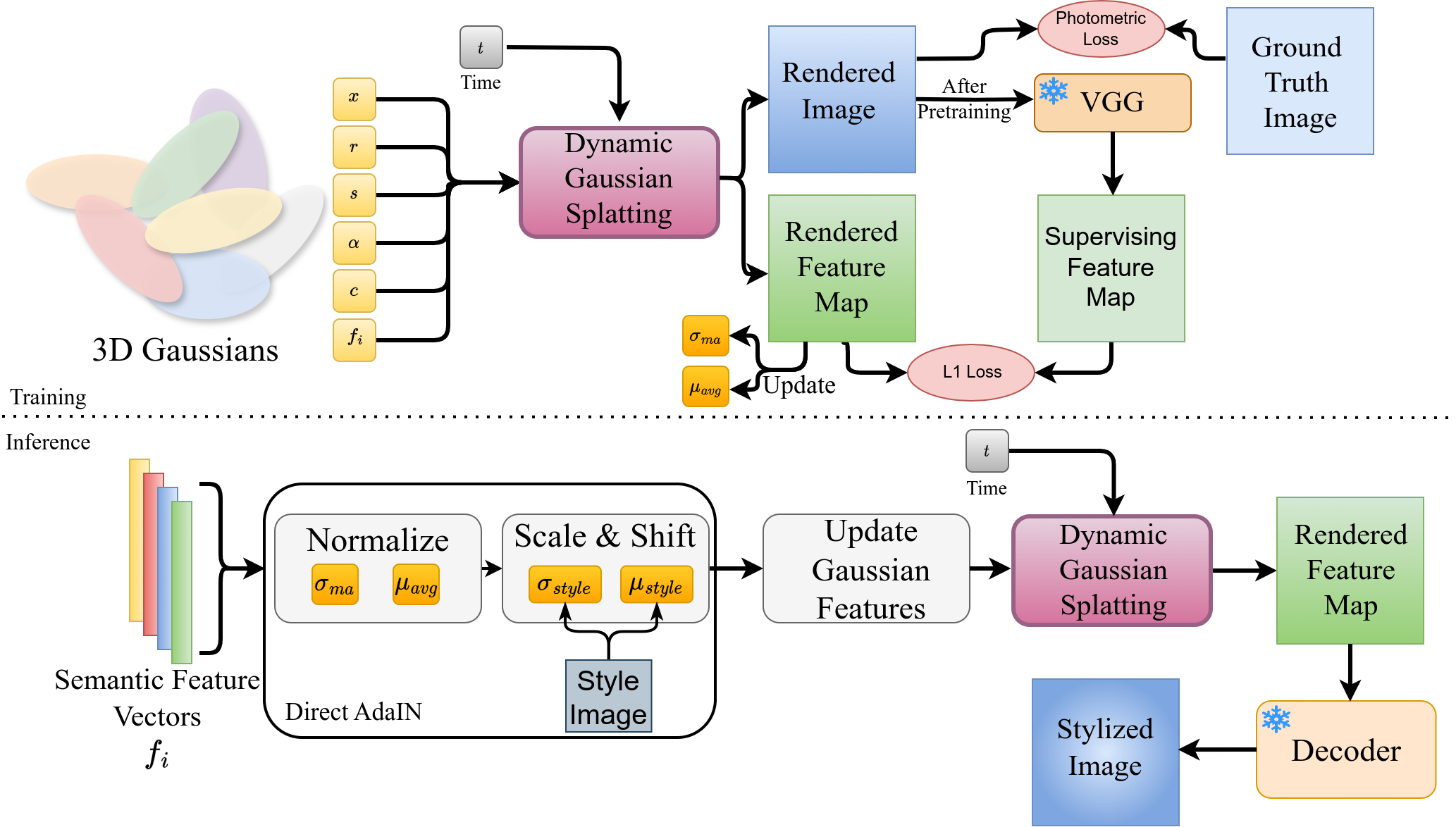}
  \caption{\textbf{Method Overview. } The above figure provides an overview of our method Z-DySS. During the training phase, we follow a straightforward pipeline, similar to ~\cite{zhou2024feature3dgs,labe2024dgd}. In addition, we compute the moving average mean and sigma of the rendered feature map that is used during the inference time to normalize the learnt semantic feature vector $f_i$ of each 3D Gaussian, before being scaled and shifted by the feature properties of the style image $S_i$. We then render these stylized feature map for a given view and timestep, before decoding to obtain the stylized novel view.}
    \label{fig:overview}
\end{figure*}

\section{Preliminaries}
\label{sec:preliminaries}

\subsection{Gaussian Splatting}

3D Gaussian Splatting(3DGS)\cite{kerbl3Dgaussians} is the latest paradigm for representing 3D scenes and performing novel view synthesis(NVS) with a fast training and inference regime.
3DGS, an explicit representation for static scenes, is made up of blob like structures, known as 3D Gaussians. Each of these Gaussians contains detailed information about itself, such as its mean position $\mu \in \mathbb{R}^{3}$ and a covariance matrix $\Sigma \in \mathbb{R}^{3 \times 3}$ as

\begin{equation}
\label{formula:gaussian's main formula}
    G(X)=e^{-\frac{1}{2}\mu^T\Sigma^{-1}\mu}.
\end{equation}

The covariance matrix $\Sigma$ is broken down into a rotation matrix $\mathbf{R}$ and a scaling matrix $\mathbf{S} \in \mathbb{R}^{3}$ as
\begin{equation}
\label{formula:sigma decom}
    \Sigma = \mathbf{R}\mathbf{S}\mathbf{S}^T\mathbf{R}^T.
\end{equation}

Each Gaussian additionally contains other learnable parameters such as its opacity $\alpha$ and color values, which are view-dependent owing to the spherical harmonics coefficients used to represent them. Furthermore, \cite{zhou2024feature3dgs} introduced having an additional semantic feature vector to each Gaussian $f \in \mathbb{R}^{N}$, of arbitrary length allowing it to distill meaningful semantic features onto the 3D feature fields.

These Gaussians, are then, projected onto the 2D image and a rendered feature map by using volumetric rendering, and the per-pixel color $C$ and feature value $F_{r}$  is given by 
\begin{equation}
C = \sum_{i \in \mathcal{N}} c_i \alpha_iT_i, \quad F_{r} = \sum_{i \in \mathcal{N}} f_i \alpha_iT_i, 
\label{eq:frender}
\end{equation}
where  $T_i$ is the transmittance, and $\mathcal{N}$ is the set of sorted Gaussians that overlap with the particular pixel \cite{zhou2024feature3dgs}. 

The parameters of Gaussians are optimized for the following loss function:
\begin{equation}
\label{eq:feature_3dgs_loss}
    \mathcal{L} = (1-\lambda)\mathcal{L}_{1} + \lambda\mathcal{L}_{D-SSIM} + \| F_{r} - F_{s}({I_{gt}}) \|_1,
\end{equation}

where $\mathcal{L}_{1}$ and $\mathcal{L}_{D-SSIM}$ are computed between the generated image $I_{gen}$ and the ground truth view $I_{gt}$. $F_{r}, F_{s}({I_{gt}})$ are the feature maps obtained via volumetric rendering of the semantic features $f_{i}$, and passing $I_{gt}$ through a pretrained foundational model respectively.

For a more in-depth explanation, we kindly refer the reader to \cite{kerbl3Dgaussians,zhou2024feature3dgs}.

However, this formulation only works for static scenes. To incorporate the moving parts, most works \cite{wu20244dgs} follow a design component where they deform the initial mean position $\mu$ of each Gaussian given a the timestamp $t$. Different approaches have been proposed to compute these deformations, either by using a small neural network \cite{yang2024deformable} or using a more sophisticated mechanism such as hexplanes \cite{wu20244dgs}. The recent work of \cite{labe2024dgd} also exhibit the effectiveness of feature distillation on such dynamic scenes.

\subsection{Adaptive Instance Normalization}
Adaptive Instance Normalization (AdaIN)\cite{huang2017adain}, since its inception has been widely used as the leading method of performing style transfer. Given a content image $C_i$ and a style image $S_i$, AdaIN matches the spatial mean and variance of the features of $C_i$ and $S_i$, namely $F_{c}$ and $F_{s}$ across each channel in the following manner:

\begin{equation}
\label{eq:adain}
\textrm{AdaIN}(F_c, F_s)= \sigma(F_s)\left(\frac{F_c-\mu(F_c)}{\sigma(F_c)}\right)+\mu(F_s)
\end{equation}

where $\mu$ and $\sigma$ are the mean and variance of the input signal respectively, and the features $F_c$ and $F_s$ are obtained by passing $C_i$ and $ S_i$ through an encoding network, such as VGG\cite{simonyan2014vgg}. The resulting feature vector after the AdaIN computation is then decoded to obtain the stylized image.


\begin{table*}[h!]
\resizebox{\textwidth}{!}{
\centering
\renewcommand{\arraystretch}{1.2} 

\begin{tabular}{lcccccccc}
\toprule
& \multicolumn{4}{c}{\textbf{Fixing camera viewpoint}} & \multicolumn{4}{c}{\textbf{Fixing time}} \\
\cmidrule(lr){2-5} \cmidrule(lr){6-9}
Method & \multicolumn{2}{c}{Short-range consistency$\downarrow$} & \multicolumn{2}{c}{Long-range consistency$\downarrow$} & \multicolumn{2}{c}{Short-range consistency$\downarrow$} & \multicolumn{2}{c}{Long-range consistency$\downarrow$} \\
\cmidrule(lr){2-3} \cmidrule(lr){4-5} \cmidrule(lr){6-7} \cmidrule(lr){8-9}
 & RMSE & LPIPS & RMSE & LPIPS & RMSE & LPIPS & RMSE & LPIPS \\
\midrule
AdaIN-4DGS  & 1.67 & 0.10 & 3.06 & 0.24 & 21.01 & 4.11 & 60.12 & 39.07 \\
4DGS-AdaIN  & 42.17 & 22.51  & 43.02 & 22.92 & 51.58 & 25.87 & 75.86 & 39.86 \\
S-Dyrf ~\cite{li2024sdyrf} & 7.23 & 1.09 & 4.43 & 0.56 & 12.46 & 1.76 & 54.31 & 22.66 \\
\textbf{Ours} & 4.82 & 0.52 & 6.46 & 0.83 & 39.64 & 13.10 & 68.67 & 38.53 \\
\bottomrule
\end{tabular}
}
\caption{\textbf{Quantitative Results} In this table, we show a quantitative benchmark of our method against the baselines. The metrics are scaled by $10^3$ for readability. The performance of ZDySS is at par with the other baselines, despite not being optimized over each style independently. It is also worth noting here that due to the excessive smoothness of the generated outputs by the synthetic baselines, they are favored strongly by the consistency metric. However, it is not always a measure of true quality as qualitatively, these methods suffer from visual artifacts as displayed in \Cref{fig:qualitative}. The metric was computed over a randomly chosen set of four diverse style images.}

\label{table:quant}
\end{table*}

\section{Method}
\label{sec:method}
Our objective is to stylize a dynamic scene given a style image $S_i$, such that the scene follows the appearance and style of $S_i$ at all timestamps and different camera positions while maintaining consistency across the spatio-temporal domain. An overview of our pipeline can be seen in \Cref{fig:overview}. 
Our method is divided into two stages. First, we train a dynamic Gaussian representation of the scene, where we enhance the Gaussians by feature vectors that are aligned with 2D VGG features. This allows us to perform zero-shot stylization at inference time based on Adaptive Instance Normalization.

\subsection{Training of the Dynamic Gaussians}
\label{sec:train_gaussians}
For the training of the dynamic Gaussian splatting, we build on top of 4DGS~\cite{wu20244dgs} framework. 4DGS leverages hexplanes~\cite{Cao2023HexPlane} to learn a deformation vector for the positions of each 3D Gaussian at any given timestamp. By design, hexplanes, and consequently 4DGS, is a fast and adaptable method for representing dynamic scenes. In addition to the usual learnable parameters for each Gaussian, we add a learnable feature vector $f_i$, similar to ~\cite{zhou2024feature3dgs, labe2024dgd}. These vectors are learnt during the the training process by rendering them onto a feature map $F_{r}$ and supervising from the ground truth feature maps from a pre-trained model. In our case, we use a pretrained VGG encoder ~\cite{simonyan2014vgg} to supervise $F_{r}$. Mathematically, it can be written as:

\begin{equation}
\mathcal{L}_f = \| F_{r} - F_{s}({I_{i}}) \|_1
\label{eq:Ll1feature}
\end{equation}

where $F_s$ is the supervising feature map signal. Therefore, the final loss function becomes the following:

\begin{equation}
\mathcal{L} = \| I_i - I_{gt} \|_1 + \| F_{r} - F_{s}({I_{i}}) \|_1
\label{eq:total_loss}
\end{equation}

We observe that pretraining the network before learning $f_i$ leads to an improvement in the final result, the details of which has been studied in the form of an ablation in \Cref{sec:ablation}. Since the motivation behind learning the rendered features from VGG stems from using a pretrained AdaIN framework, we keep a track of the moving average mean $\mu_{avg}$ and moving average standard deviation $\sigma_{ma}$ of the rendered features $F_{r}$. This is done to mitigate the effect of normalizing $F_r$ for each view, which in turn is a possible cause of consistencies. It is worth noting here that, unlike most of the previous methods, we have not used any style images in the training process. 

\subsection{Inference}
One of the naive ways to perform zero-shot stylization is to perform AdaIN operation on the rendered feature map $F_r$ with the given style image $S_i$, followed by decoding the resulting feature map into the stylized novel view. This is a potential source of inconsistency amongst multiple views, as the rendered feature maps are each normalized independently of each other. Therefore, we use the moving average mean $\mu_{avg}$, and standard deviation $\sigma_{ma}$ computed during the training process for normalizing the feature maps. To avoid redundant computation, due to the linearity of the affine operation, as shown in ~\Cref{eq:adain} and ~\Cref{eq:frender}, we perform the AdaIN directly on the learnt gaussian feature vectors $f_i$. We show that this helps us maintain superior performance than the naive method described above and has been studied further in the ablations section. Once $f_i$ is carrying the style information, we render them into a feature map, which are then directly fed into a pretrained decoder from ~\cite{huang2017adain} to obtain the stylized image.

\begin{figure*}[h]
    \centering
    \includegraphics[width=\textwidth]{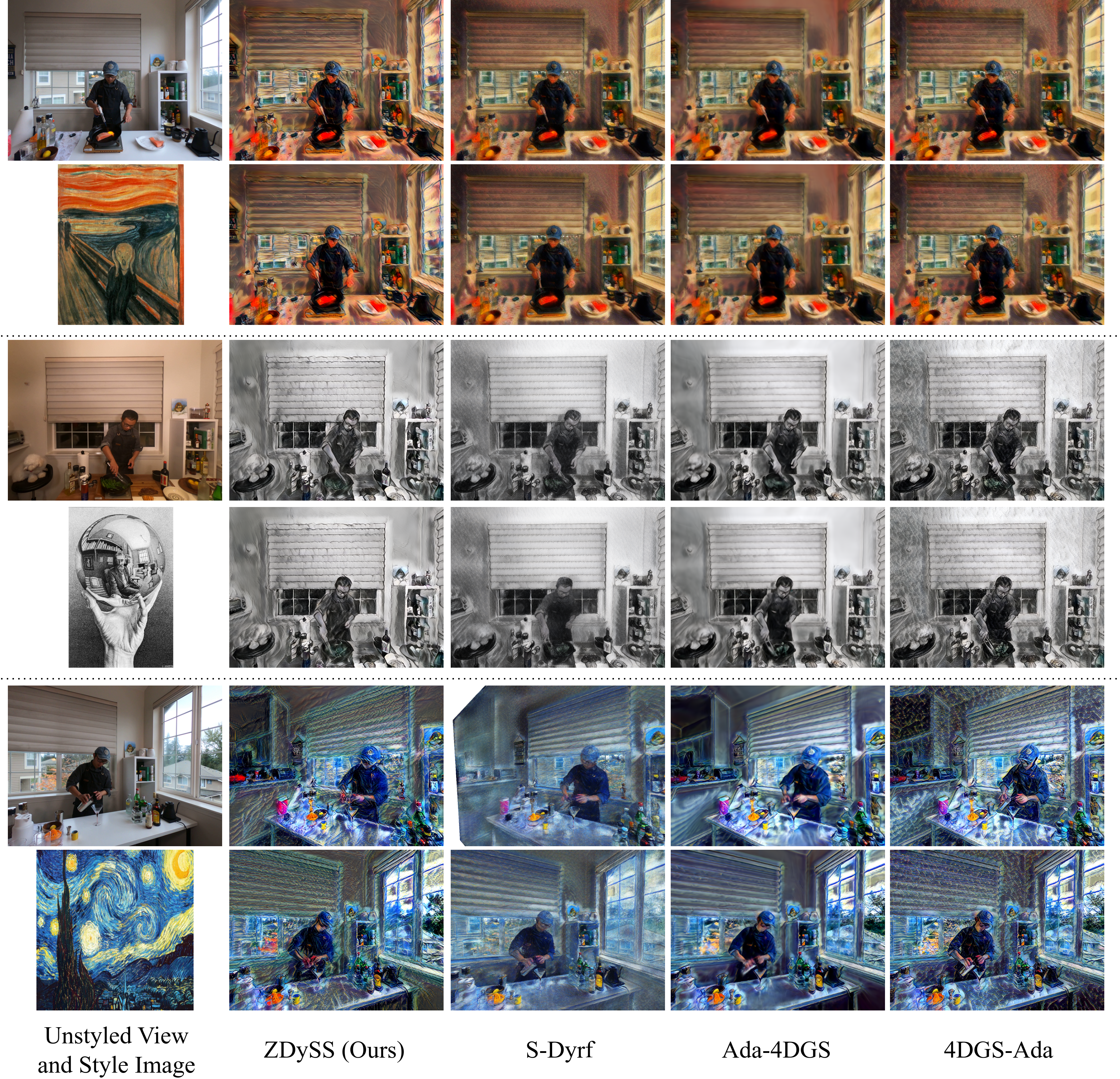}
  \caption{\textbf{Qualitative Results} Here we show a comparative study of ZDySS against the baselines, namely S-DyRF, Ada-4DGS, and 4DGS-Ada. It can be observed here that, despite not being optimized on every queried style image, ZDySS is able to faithfully stylize the given scene at various timesteps and viewpoints. ZDySS also retains most details out of all the methods, while carrying the style information. For instance, Ada-4DGS and 4DGS-Ada suffer from the problem of having spikey and elongated Gaussians, along with strong blurriness, especially along the high frequency regions. S-Dyrf on the other hand, suffers from blurriness as compared to our method. In addition, we also provide videos in the supplementary that are effective in displaying the consistency differences between ours and the mentioned baselines.}
    \label{fig:qualitative}
\end{figure*}

\begin{table*}[h!]
\resizebox{\textwidth}{!}{
\centering
\renewcommand{\arraystretch}{1.2} 

\begin{tabular}{lcccccccc}
\toprule
& \multicolumn{4}{c}{\textbf{Fixing camera viewpoint}} & \multicolumn{4}{c}{\textbf{Fixing time}} \\
\cmidrule(lr){2-5} \cmidrule(lr){6-9}
Method & \multicolumn{2}{c}{Short-range consistency$\downarrow$} & \multicolumn{2}{c}{Long-range consistency$\downarrow$} & \multicolumn{2}{c}{Short-range consistency$\downarrow$} & \multicolumn{2}{c}{Long-range consistency$\downarrow$} \\
\cmidrule(lr){2-3} \cmidrule(lr){4-5} \cmidrule(lr){6-7} \cmidrule(lr){8-9}
 & RMSE & LPIPS & RMSE & LPIPS & RMSE & LPIPS & RMSE & LPIPS \\
\midrule
Naive ZDySS & 4.87 & 0.54 & 6.57 & 0.88 & 41.09 & 13.52 & 68.78 & 37.37 \\
\textbf{Ours} & 4.82 & 0.52 & 6.46 & 0.83 & 39.64 & 13.10 & 68.67 & 38.53 \\
\bottomrule
\end{tabular}
}
\caption{\textbf{Quantitative Results: Naive Ablation} We show here the effect of using our running mean against normalizing using the mean and standard deviation of each rendered feature map individually. We can see that using our method is more consistent as it increases spatio-temporal consistency.}

\label{table:naive}
\end{table*}

\section{Experiments}

\subsection{Implementation Details}
Our framework is built on top of the implementation of ~\cite{wu20244dgs}. We pretrain the framework for 14000 iterations, followed by the joint training of the semantic features and other Gaussian parameters for 7000 iterations. The length of our semantic feature vector $f_i$ per Gaussian is 512. For the rendering process, we adopt the renderer from ~\cite{zhou2024feature3dgs}, that renders a feature map of 512 channel dimension. We allow the Gaussians to densify and prune in the entire process of 21000 iterations. For the encoding of the images into latent space, we use a pre-trained VGG encoder \cite{simonyan2014vgg}, followed by a pre-trained decoder in ~\cite{huang2017adain}, both of which are kept frozen during the entire pipeline. 

\subsection{Datasets and Baselines}
For the purpose of our experiments, we chose the real-world Plenoptic Video Dataset~\cite{li2022neural3dvideo}. The dataset contains 6 high quality scenes taken by 20 synchronized cameras for 10 seconds at 30FPS. Since the task of dynamic scene stylization is relatively new, there are not many established baselines that are available. We use S-DyRF~\cite{li2024sdyrf} as one of the baselines for comparison to our method. Even though we do not require an optimization over each style image, we evaluate against S-DyRF for the sake of completeness. In addition, we create two synthetic baselines, namely Ada-4DGS and 4DGS-Ada. These baselines are based upon the Gaussian Splatting framework. In Ada-4DGS, we train a 4D Gaussian splatting scene trained on stylized ground truth images, whereas in 4DGS-Ada, we apply AdaIN on the rendered image for each view separately. Ada-4DGS also an example of the "overfit" method like S-DyRF since they cannot handle unseen styles at inference time without re-training. 

\subsection{Qualitative Results}
We provide qualitative results in ~\Cref{fig:qualitative}. As we can observe, ZDySS provides high quality stylized novel views while maintaining the scene properties and consistency. On a closer look, we can see the missing details in the synthetic baselines, namely Ada-4DGS and 4DGS-Ada. That is because, in Ada-4DGS, the 4DGS process smoothens out all the inconsistent details, and hence, there are missing details. Also, Ada-4DGS suffers from spiking Gaussians, that happens due to the optimization process not being able to cover the high frequency areas with the required number of Gaussians. We on the other hand, do not suffer from this problem due to the use of a decoder. 

For a more comprehensive evaluation, we provide videos of all the methods in the supplementary material.

\subsection{Quantitative Results}
We demonstrate the effectiveness of our method in a quantitative comparison in ~\Cref{table:quant}. Following prior stylization works, we adopt the view consistency metric as a measure for model performance. As shown in ~\cite{li2024sdyrf}, since we are deadling with dynamic scenes, it is only natural to also perform the consistency measure not only across multiple views, but also across the temporal domain, keeping the camera viewpoint fixed. Following prior works, we measure consistency by using a pretrained RAFT~\cite{teed2020raft} to warp one view into a reference view, and simply computing a RMSE and LPIPS distance between the warped view and the reference view. Mathematically, it can be shown as:

\begin{equation}
    \mathbb{E}_{wlpips}(\mathcal{O}_{v},\mathcal{O}_{v^{'}}) = LPIPS(\mathcal{O}_{v}, \mathcal{M}_{v}(\mathcal{W}(\mathcal{O}_{v^{'}})) )
\end{equation}
and
\begin{equation}
    \mathbb{E}_{wrmse}(\mathcal{O}_{v},\mathcal{O}_{v^{'}}) = RMSE(\mathcal{O}_{v}, \mathcal{M}_{v}(\mathcal{W}(\mathcal{O}_{v^{'}})) ),
\end{equation}

where $\mathcal{W}$ , $\mathcal{O}_{v}$, $\mathcal{M}_{v}$, , and are the warping, rendered view and the maskingrespectively for two views ${v}$ and ${v}^{'}$.

The numbers were computed from a randomly chosen set of four style images. The metric is, by design, more favourable to images that are smoothened out or blurry, which is not a desired output in term of visual quality, and hence we see the metrics low for the synthetic baselines while having minimal visual quality.

\begin{figure*}[h]
    \centering
    \includegraphics[width=\textwidth]{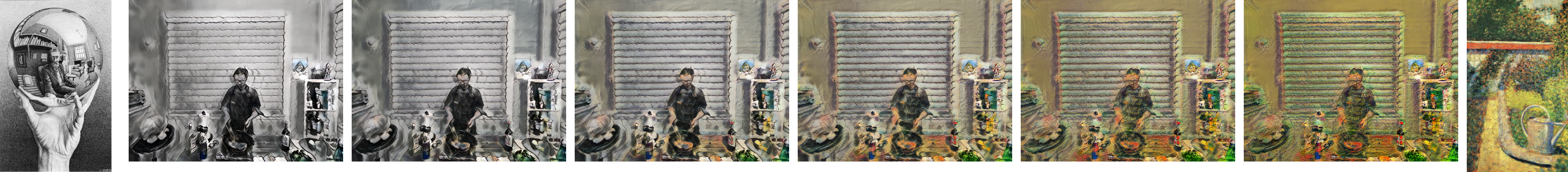}
  \caption{\textbf{Style Interpolation} We interpolate between the latent vectors of two different style images at test time, obtaining meaningful stylizations as we move from one style to another.}
    \label{fig:interp}
\end{figure*}

\begin{figure}[h]
    \centering
    \includegraphics[width=0.5\textwidth]{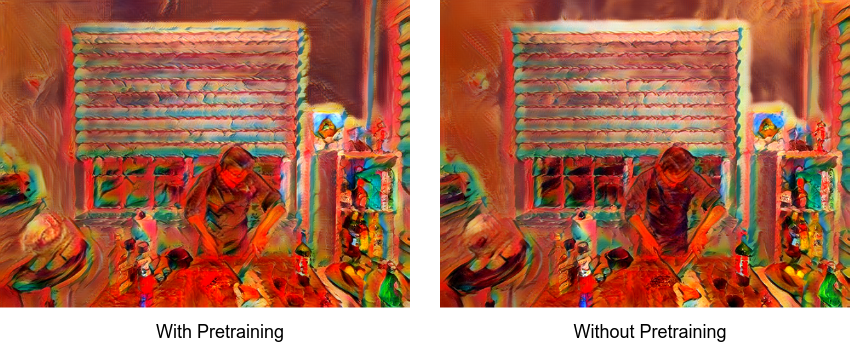}
  \caption{\textbf{Pretraining} Pretraining the scene initially without the feature map supervision helps retain finer details in the stylized outputs. We pretrain the scene for 14000 iterations, as suggested in ~\cite{wu20244dgs}.}
    \label{fig:pretraining}
\end{figure}

\section{Ablations}
\label{sec:ablation}

\subsection{Style Interpolation}
In this ablation, we interpolate between the latent vectors of the style images at inference time. It shows that not only we are able to faithfully stylize the input scene, the method is able to meaningfully handle complex latent codes, that are formed by combining style images, thus proving the zero-shot capabilities of ZDySS.

\subsection{Effect of pre-training}
We ablate the effect of pre-training a 4DGS scene on our pipeline. We observe that pre-training helps retain finer details in the stylized image, such as the prints and patterns in the background.

\subsection{Effect of Running mean and standard deviation}
We show the effect of using our moving average mean against individual feature maps in ~\Cref{table:naive}. We can see that in both, temporal and the spatial domain, using a meaningful value for normalizing all Gaussians before mixing with the style features not only adds to the consistency, but also reduces redundant computation. 
\section{Limitations}
While our method is able to fulfill the task at hand, it does come with its set of limitations. Since we rely on pretrained models for performing zero-shot stylization, we have less control of the stylization. Also, using a decoder reduces the rendering speed of the 4DGS pipeline, improving upon which is an improvement to follow while keeping the flexibility and stylization power intact.

\section{Conclusion}
We have introduced ZDySS, a novel approach to zero-shot stylization of dynamic scenes based on Gaussian splatting. By augmenting the Gaussians with feature vectors, which we align with 2D VGG features, we can adopt Adaptive Instance Normalization for the dynamic scene stylization. We use a running average to ensure temporal and multi-view consistency of the content normalization parameters. Our results show compelling stylizations of dynamic scenes in multiple varying styles.

\newpage
{
    \small
    \bibliographystyle{ieeenat_fullname}
    \bibliography{main}
}


\end{document}